\documentclass[a4paper,english,oneside,reqno]{article}
\usepackage[T1]{fontenc}
\usepackage[latin9]{inputenc}
\usepackage{amsthm}
\usepackage{amsmath}
\usepackage{graphicx}
\usepackage{setspace}
\usepackage{amssymb}
\makeatletter
\theoremstyle{plain}
\theoremstyle{plain}
\newtheorem{thm}{Theorem}
  \theoremstyle{remark}
  \newtheorem{rem}[thm]{Remark}
 \theoremstyle{definition}
  \newtheorem{example}[thm]{Example}

\usepackage[colorlinks=true, pdfstartview=FitV, linkcolor=blue,citecolor=blue, urlcolor=blue]{hyperref}
\usepackage[numbers,sort]{natbib}

\makeatother

\usepackage{babel}

\begin{document}

\title{Descriptive-complexity based distance for fuzzy sets}

\author{Laszlo Kovacs%
\thanks{Department of Information Technology, University of Miskolc, HUNGARY.
H-3515. Miskolc-Egyetemvaros. \emph{Email}: kovacs@iit.uni-miskolc.hu %
} ~and Joel Ratsaby%
\thanks{Department of Electrical and Electronics Engineering, Ariel University
Center of Samaria, Ariel 40700, ISRAEL. \emph{Email}: ratsaby@ariel.ac.il.
\emph{(Corresponding author}).%
}}
\maketitle
\begin{abstract}
A new distance function $\text{dist}\left(A,B\right)$ for fuzzy sets
$A$ and $B$ is introduced. It is based on the descriptive complexity,
i.e., the number of bits (on average) that are needed to describe
an element in the symmetric difference of the two sets. The distance
gives the amount of additional information needed to describe any
one of the two sets given the other. We prove its mathematical properties
and perform pattern clustering on data based on this distance.
\end{abstract}
\vspace{0.4cm}

\emph{Key words}: Fuzzy sets, descriptive complexity, entropy, distance.

\section{\label{sec:Introduction}Introduction}

The notion of distance between two objects is very general. Distance
metrics and distances have now become an essential tool in many areas
of mathematics and its applications including geometry, probability,
statistics, coding/graph theory, data analysis, pattern recognition.
For a comprehensive source on this subject see \cite{Deza-09}. The
notion of a fuzzy set was introduced by \cite{Zadeh65}. It is a class
of objects with continuous values of membership and hence extends
the classical definition of a set (to distinguish it from a fuzzy
set we refer to it as a crisp set). Formally, a fuzzy set is a pair
$\left(E,m\right)$ where $E$ is a set of objects and $m$ is a membership
function $m:E\rightarrow\left[0,1\right]$. Fuzzy set theory can be
used in a wide range of domains in which information is incomplete
or imprecise, such as pattern recognition, decision theory. The concept
of distance and similarity is important in the area of fuzzy logic
and sets. We now review some common ways of defining distances on
fuzzy sets (see \cite{Zwick87} and references therein). 

Classical distances measure how far two points are in Euclidean space.
For instance, the Minkowski distance between two points $x$ and $y$
in $\mathbb{R}^{n}$ is defined as\begin{equation}
d_{r}(x,y):=\left(\sum_{i=1}^{n}\left|x_{i}-y_{i}\right|^{r}\right)^{1/r},\quad r\geq1.\label{eq:minkow}\end{equation}
Let $E$ be a finite set and let $\Phi(E)$ be the set of all fuzzy
subsets of $E$. Consider $A$,$B$ two fuzzy subsets $A$, $B\in\Phi(E)$
with membership functions $m_{A}$,$m_{B}:E\rightarrow\left[0,1\right]$.
Then (\ref{eq:minkow}) can be extended to the following distance,
\[
d_{r}(A,B):=\left(\sum_{x\in E}\left|m_{A}\left(x\right)-m_{B}(x)\right|^{r}\right)^{1/r},\quad r\geq1.\]
Based on (\ref{eq:minkow}) letting $r=2$ we have the Hausdorff distance
between two non-empty compact crisp sets $U$, $V\subset\mathbb{R}$,
\begin{equation}
q\left(U,V\right):=\max\left\{ \sup_{v\in V}\inf_{u\in U}d_{2}\left(u,v\right),\sup_{u\in U}\inf_{v\in V}d_{2}\left(u,v\right)\right\} .\label{eq:quv}\end{equation}
 This can be extended to fuzzy sets as follows: let $A\in\Phi(E)$
be a fuzzy set and denote by $A_{\alpha}$ the $\alpha$-level set
of the fuzzy set $A$ which is defined as $A_{\alpha}=\left\{ x\in E\mid m_{A}(x)\geq\alpha\right\} $.
Then for two fuzzy subsets $A$, $B\in\Phi(E)$ the distance in (\ref{eq:quv})
can be extended to the following distance between $A$ and $B$,\[
q(A,B):=\int_{0}^{1}q(A_{\alpha},B_{\alpha})d\alpha.\]
Another approach is based on set-theoretic distance functions. For
a fuzzy set $A\in\Phi(E)$ define the cardinality of $A$ as $\left|A\right|=\sum_{x\in E}m_{A}(x)$.
Extend the intersection and union operations by defining the membership
functions\[
m_{A\cap B}(x):=\min\left\{ m_{A}(x),m_{B}(x)\right\} \]
and 

\[
m_{A\cup B}(x):=\max\left\{ m_{A}(x),m_{B}(x)\right\} .\]
Then for fuzzy sets $A$, $B\in\Phi(E)$ we may define the distance
function\[
S_{1}(A,B):=1-\frac{\left|A\cap B\right|}{\left|A\cup B\right|}=1-\frac{\sum_{x\in E}m_{A\cap B}(x)}{\sum_{x\in E}m_{A\cup B}(x)}.\]
Another distance is based on four features of a fuzzy set. Let the
domain of interest be $\mathbb{R}$ and consider a fuzzy set $A$
in $\Phi(\mathbb{R})$. The power of $A$ (which extends the notion
of cardinality) is defined as \[
\text{power}\left(A\right):=\int_{-\infty}^{\infty}m_{A}(x)dx.\]
 Let $S(x)=-x\ln x-(1-x)\ln(1-x)$ then define the entropy of $A$
as \[
\text{entropy}\left(A\right):=\int_{-\infty}^{\infty}S(m_{A}(x))dx.\]
Define the centroid as \[
c(A):=\frac{\int_{-\infty}^{\infty}xm_{A}(x)dx}{\text{power}\left(A\right)}\]
 and the skewness as \[
\text{skew}\left(A\right):=\int_{-\infty}^{\infty}\left(x-c(A)\right)^{3}m_{A}(x)dx.\]
Let $v(A)=\left[\text{power}\left(A\right),\text{ entropy}\left(A\right),c(A),\text{ skew}(A)\right]$
then \cite{Bonisson79} defines the distance between two fuzzy sets
$A$, $B\in\Phi(\mathbb{R})$ as the Euclidean distance $\left\Vert v(A)-v(B)\right\Vert $.

We now proceed to discuss the notion of distances that are based on
descriptive complexity of sets.

\section{Information based distances}

A good distance is one which picks out only the `true' dissimilarities
and ignores factors that arise from irrelevant variables or due to
unimportant random fluctuations that enter the measurements. In most
applications the design of a good distance requires inside information
about the domain, for instance, in the field of information retrieval
\cite{YAtes99} the distance between two documents is weighted largely
by words that appear less frequently since the words which appear
more frequently are less informative. 

Typically, different domains require the design of different distance
functions which take such specific prior knowledge into account. It
can therefore be an expensive process to acquire expertise in order
to formulate a good distance. Recently, a new distance for sets was
introduced \cite{RatsabyMATCOS2010} which is based on the concept
of descriptional complexity (or discrete entropy). A description of
an object in a finite set can be represented as a finite binary string
which provides a unique index of the object in the set. The description
complexity of the object is the minimal length of a string that describes
the object. The distance of \cite{RatsabyMATCOS2010} is based on
the idea that two sets should be considered similar if given the knowledge
of one the additional complexity in describing an element of the other
set is small (this is also referred to as the conditional combinatorial
entropy, see \cite{RATSABY_DBLP:conf/sofsem/Ratsaby07,Ratsaby_IW}
and references therein). The advantage in this formulation of distance
is its universality, i.e., it can be applied without any prior knowledge
or assumption about the domain of interest, i.e., the elements that
the sets contain. Such a distance can be viewed as an information-based
distance since the conditional descriptional complexity is essentially
the amount of information needed to describe an element in one set
given that we know the other set (for more on the notion of combinatorial
information and entropy see \cite{RATSABY_DBLP:conf/sofsem/Ratsaby07,Ratsaby_IW}).

In the current paper we introduce a distance function between two
general sets, i.e., sets that can be crisp or fuzzy. Following the
information-based approach of \cite{RatsabyMATCOS2010} we resort
to entropy as the main operator that gives the measure of dissimilarity
between two sets. We use the membership of the symmetric difference
of two sets as the probability of a Bernoulli random variable whose
entropy is the expected description complexity of an element that
belongs to only one of the two sets. Thus the distance function measures
how many bits (on average) are needed to describe an element in the
symmetric difference of the two sets. In other words, it is the amount
of additional information needed to describe any one of the two sets
given knowledge of the other.

Being a description-complexity based distance gives it certain characteristic
properties. For instance, the distance between a crisp set $A$ and
its complement $\overline{A}$ is zero since there is no need for
additional information in order to describe one of these two sets
when knowing the other. That is, knowledge of a set $A$ automatically
implies knowing the set $\overline{A}$ and vice versa. They are clearly
not equal but our distance function cleverly renders them as the most
similar that two sets can be (zero distance apart).

The next section formally introduces the distance and in Theorem \ref{thm:The-function-}
we prove its metric properties.

\section{Distance function}

We write \emph{w.p.} for {}``with probability''. Let $[N]=\left\{ 1,\ldots,N\right\} $
be a domain of interest. Let $A\in\Phi([N])$ be a set with membership
function $m_{A}:[N]\rightarrow[0,1]$. We use $x$ to denote a value
in $[N]$. Given two fuzzy subsets $A$, $B\in\Phi([N])$ with membership
functions $m_{A}(x)$, $m_{B}(x)$, as mentioned in section \ref{sec:Introduction}
we denote by \[
m_{A\cup B}(x):=\max\left\{ m_{A}(x),m_{B}(x)\right\} \]
 and 

\[
m_{A\cap B}(x):=\min\left\{ m_{A}(x),m_{B}(x)\right\} .\]
Define by $A\vartriangle B=(A\bigcup B)\setminus(A\bigcap B)$ the
symmetric difference between crisp sets $A$,$B$. For fuzzy sets
$A$, $B\in\Phi([N])$ define by \[
m_{A\vartriangle B}(x):=m_{A\cup B}(x)-m_{A\cap B}(x).\]
Define a sequence of Bernoulli random variables $X_{A}(x)$ for $x\in[N]$
taking the value $1$ w.p. $m_{A}(x)$ and the value $0$ w.p. $1-m_{A}(x)$.
Define by $H(X_{A}(x))$ the entropy of $X_{A}(x)$,\[
H(X_{A}(x)):=-m_{A}(x)\log m_{A}(x)-(1-m_{A}(x))\log(1-m_{A}(x)).\]
Define the random variable

\[
X_{A\vartriangle B}(x):=\left\{ \begin{array}{ccc}
1 & w.p. & m_{A\vartriangle B}(x)\\
0 & w.p. & \;\;1-m_{A\vartriangle B}(x).\end{array}\right.\]
We define a new distance between $A$, $B\in\Phi([N])$ as \[
\text{dist}(A,B):=\frac{1}{N}\sum_{x=1}^{N}H(X_{A\vartriangle B}(x))\]

\begin{rem}
This definition can easily be extended to the case of an infinite
domain, for instance, a subset of the real line. In that case the
distance can be defined as the expected value of $\mathbb{E}H(X_{A\vartriangle B}(\xi))$
where $\xi$ is a random variable with some probability distribution
$P(\xi)$ with respect to which the expectation is computed. 
\end{rem}
The next theorem shows that the distance satisfies the metric properties.
\begin{thm}
\label{thm:The-function-}The function $\text{dist}(A,B)$ is a semi-metric
on $\Phi([N])$, i.e., it is non-negative, symmetric, it equals zero
if $A=B$ and it satisfies the triangle inequality. \end{thm}
\begin{rem}
Note that the function $\text{dist}(A,B)$ may equal zero even when
$A\neq B$. 

We now prove Theorem \ref{thm:The-function-}.\end{rem}
\begin{proof}
Since the entropy function is non-negative (see for instance, \cite{CoverThomas2006})
then for any two subsets $A$, $B\in\Phi([N])$ we have $\text{dist}(A,B)\geq0$.
It is easy to see that the symmetry property is satisfied since for
every $x\in[N]$ we have $X_{A\vartriangle B}(x)=X_{B\vartriangle A}(x)$.
For every subset $A$ the value $\text{dist}(A,A)=0$ since $m_{A\vartriangle A}(x)=0$
hence $H(X_{A\vartriangle A}(x))=0$ for all $x$.

Let us now show that the triangle inequality holds. Let $A$, $B$,
$C$ be any three elements of $\Phi([N])$. Fix any point $x\in[N]$
and without loss of generality suppose that $m_{A}(x)\leq m_{B}(x)\leq m_{C}(x)$.
Denote by $p=m_{B}(x)-m_{A}(x)$ and $q=m_{C}(x)-m_{B}(x)$. Without
loss of generality assume that $p\leq q$. Then we have $m_{A\vartriangle C}(x)=p+q$.
Denote by $H(p)$, $H(q)$ and $H(p+q)$ the entropies $H(X_{A\vartriangle B}(x))$,
$H(X_{B\vartriangle C}(x))$ and $H(X_{A\vartriangle C}(x))$ respectively.
We aim to show that $H(p+q)\leq H(p)+H(q).$ This will imply that
for every $x\in[N]$, $H(X_{A\vartriangle C}(x))\leq H(X_{A\vartriangle B}(x))+H(X_{B\vartriangle C}(x))$
and hence it holds for the average $\frac{1}{N}\sum_{x}H(X_{A\vartriangle C}(x))\leq\frac{1}{N}\sum_{x}H(X_{A\vartriangle B}(x))+\frac{1}{N}\sum_{x}H(X_{B\vartriangle C}(x))$.

We start by considering the straight line function $\ell:[0,1]\rightarrow[0,1]$
defined as: \[
\ell(z):=\frac{H(q)-H(p)}{q-p}z+H(p)-\frac{H(q)-H(p)}{q-p}p\]
 which cuts through the points $(z,\ell(z))=(p,H(p))$ and $(z,\ell(z))=(q,H(q))$.
For a function $f$ let $f'$ denote its derivative. We claim the
following,

\emph{Claim 1. $H'(z)\leq\ell'(z)$ for all $z\in[q,1]$.}

\emph{Proof}: The derivative of $H(z)$ is $H'(z)=\log\left(\frac{1-z}{z}\right)$.
This is a decreasing function on $[0,1]$ hence it suffices to show
that $H'(q)\leq\ell'(z)$ for all $z\in[q,1]$. The derivative of
$\ell(z)$ is $\frac{H(q)-H(p)}{q-p}$. So it suffices to show that
\[
\log\left(\frac{1-q}{q}\right)\leq\frac{H(q)-H(p)}{q-p}.\]
 This is equivalent to \begin{equation}
(q-p)\log\left(\frac{1-q}{q}\right)\leq H(q)-H(p).\label{eq:qmp}\end{equation}
The left hand side of (\ref{eq:qmp}) can be reduced to,\begin{equation}
q\log(1-q)-q\log q-p\log(1-q)+p\log q.\label{eq:qmp1}\end{equation}
 Adding and subtracting the term $(1-q)\log(1-q)$ and using $H(q)=-q\log q-(1-q)\log(1-q)$
makes (\ref{eq:qmp1}) be expressed as

\[
H(q)+p\log q+(1-p)\log(1-q).\]
Substituting this for the left hand side of (\ref{eq:qmp}) and canceling
$H(q)$ on both sides gives the following inequality which we need
to prove \[
p\log q+(1-p)\log(1-q)\leq p\log p+(1-p)\log(1-p).\]
 It suffices to show that,\begin{equation}
p\log\left(\frac{p}{q}\right)+(1-p)\log\left(\frac{1-p}{1-q}\right)\geq0.\label{eq:plp}\end{equation}
 That (\ref{eq:plp}) holds follows from the information inequality
(see Theorem 2.6.3 of \cite{CoverThomas2006}) which lower bounds
the divergence $D(P||Q)\geq0$ where $P$,$Q$ are two probability
functions and $D(P||Q)=\sum_{x}P(x)\log\frac{P(x)}{Q(x)}$. Hence
the claim is proved. $\blacksquare$

Next we claim the following:

\emph{Claim 2}. $H(p+q)\leq\ell(p+q)$.

\emph{Proof}: Consider the case that $p\leq q\leq\frac{1}{2}$. Since
$q\leq\frac{1}{2}$ then $H'(z)$ evaluated at $z=q$ is non-negative.
Hence both $H(z)$ and $\ell(z)$ are monotone increasing on $q\leq z\leq\frac{1}{2}$
and $H(q)=\ell(q)$. By Claim 1, $\ell$ increases faster than $H$
on $[q,1]$, in particular on the interval $q\leq z\leq\frac{1}{2}$.
Hence, for all $z\in[q,\frac{1}{2}]$ we have $H(z)\leq\ell(z)$.
Now, if $p+q\in[q,\frac{1}{2}]$ then it follows that $H(p+q)\leq\ell(p+q)$.
Otherwise it must hold that $p+q\in(\frac{1}{2},1]$. But $H$ is
decreasing and $\ell$ is increasing over this interval. Hence we
have $\ell(z)>\ell(\frac{1}{2})\geq H(\frac{1}{2})>H(z)$ for $z\in(\frac{1}{2},1]$,
in particular for $z=p+q$ hence $\ell(p+q)\geq H(p+q)$. This proves
the claim. $\blacksquare$

From Claim 2 it follows that \begin{eqnarray}
H(p+q) & \leq & \frac{H(q)-H(p)}{q-p}(p+q)+H(p)-\frac{H(q)-H(p)}{q-p}p\nonumber \\
 & = & H(p)+q\frac{H(q)-H(p)}{q-p}.\label{eq:sdd}\end{eqnarray}
 It suffices to show that \[
q\frac{H(q)-H(p)}{q-p}\leq H(q)\]
or equivalently, \begin{equation}
\frac{H(q)}{q}\leq\frac{H(p)}{p}.\label{eq:fp}\end{equation}
 Letting $f(z)=\frac{H(z)}{z}$ and differentiating we obtain\[
f'(z)=\frac{\log(1-z)}{z^{2}}\]
 which is non-positive for $z\in[0,1].$ Hence $f$ is non-increasing
over this interval. Since by assumption $q\geq p$ then it follows
that $f(q)\leq f(p)$ and \ref{eq:fp} holds. This completes the proof
of the theorem.
\end{proof}

\section{\label{sec:Simple-Examples}Simple Examples}

Let us evaluate this distance for a few examples. Consider two sets
$A$ and its complement $\overline{A}$. Their membership functions
satisfy the relation: \[
m_{\overline{A}}(x)=1-m_{A}(x)\]
 hence the membership function for the symmetric difference is\[
m_{A\vartriangle\overline{A}}(x)=\max\left\{ m_{A}(x),(1-m_{A}(x))\right\} -\min\left\{ m_{A}(x),(1-m_{A}(x))\right\} .\]
Note that for any $x\in[N]$ with a crisp membership value, i.e.,
$m_{A}(x)=1$, or $m_{A}(x)=0$, we have $m_{A\vartriangle\overline{A}}(x)=1$
and hence in this case $H(X_{A\vartriangle\overline{A}}(x))=0$. This
means that for a crisp set $A$ (for all $x\in A$, $m_{A}(x)\in\left\{ 0,1\right\} $)
our distance has the following property (we call this the \emph{complement-property}):
\[
\text{dist}(A,\overline{A})=0.\]
From an information theoretic perspective, this property is expected
since knowing a set $A$ automatically means that we also know how
to describe its complement. Hence there is no additional description
necessary to describe $\overline{A}$ given $A$. This is what $\text{dist}(A,\overline{A})=0$
means.

Let us now consider some examples of pairs of fuzzy sets and their
distances. Let $N=20$ and the domain be \emph{$[N]=\left\{ 1,2,\ldots,20\right\} $.
}In the following examples we plot the membership functions of several
fuzzy sets. Note, we connect the point values of the membership function
by lines in order to make the plots clearer (remember that the actual
membership functions are defined only on the discrete set $[N]$).
\begin{example}
\emph{\label{exa:Let--and} Consider the fuzzy sets $A$,$B$,$C$
and the complement $A^{c}$ with membership functions as shown in
Figure \ref{fig:Fuzzy-sets-,,}. Note, that $A$ and its complement
are crisp sets. The distance matrix $D=\left[d_{i,j}\right]$ is shown
in (\ref{eq:dist-matrix}); the rows and columns correspond to $A$,
$B$, $C$ and $A^{c}$ so that for instance the element $d_{2,3}=\text{dist}\left(B,C\right)=0.709$.
As can be seen, $C$ is a translated version of $B$ and they are
both the same distance from $A$. This is due to $H(X_{A\vartriangle B}(x))=H(X_{A\vartriangle C}(x+10))$.
$B$ and $C$ are farther apart than $B$ and $A$. Since $\text{dist}\left(A,A^{c}\right)=0$
then each one of $B$, $C$ is of the same distance to $A$ as to
$A^{c}$.}
\end{example}
\begin{equation}
D=\left(\begin{array}{cccc}
0 & 0.354 & 0.354 & 0\\
0.354 & 0 & 0.709 & 0.354\\
0.354 & 0.709 & 0 & 0.354\\
0 & 0.354 & 0.354 & 0\end{array}\right)\label{eq:dist-matrix}\end{equation}

\begin{example}
\emph{Continuing with the same domain as in Example \ref{exa:Let--and}
let us consider the fuzzy sets $A$,$B$,$C$ and the complement $A^{c}$
with membership functions as shown in Figure \ref{fig:Fuzzy-sets-,,-1}.
The membership function of the set $C$ is now flat and as the distance
matrix $D=\left[d_{i,j}\right]$ in (\ref{eq:dist-1}) shows $C$
is now farther from $B$ (which has a triangular membership function).
As in the previous example $B$ remains closer to $A$ than to $C$. }
\end{example}
\begin{equation}
D=\left(\begin{array}{cccc}
0 & 0.354 & 0.5 & 0\\
0.354 & 0 & 0.854 & 0.354\\
0.5 & 0.854 & 0 & 0.5\\
0 & 0.354 & 0.5 & 0\end{array}\right)\label{eq:dist-1}\end{equation}

\begin{example}
\emph{Continuing with the same domain as in Example \ref{exa:Let--and}
let us consider the fuzzy sets $A$,$B$,$C$ and the complement $A^{c}$
with membership functions as shown in Figure \ref{fig:Fuzzy-sets-,,-1-1}.
Note that now $C$ is translated from $B$ by an amount that is smaller
compared to Example \ref{exa:Let--and}. As can be seen from the distance
matrix of (\ref{eq:dist-2}) this results in a smaller distance $\text{dist}\left(B,C\right)=0.336$
compared to $0.709$. As in Example \ref{exa:Let--and} , $A$ is
as similar to $B$ as to $C$ since the distance $\text{dist}\left(A,B\right)=\text{dist}\left(A,C\right)=0.354$. }
\end{example}
\begin{equation}
D=\left(\begin{array}{cccc}
0 & 0.354 & 0.354 & 0\\
0.354 & 0 & 0.336 & 0.354\\
0.354 & 0.336 & 0 & 0.354\\
0 & 0.354 & 0.354 & 0\end{array}\right)\label{eq:dist-2}\end{equation}

\section{Clustering using the distance}

We tested the proposed distance function on real data. The data%
\footnote{The data set is the European Social Survey Round 4 Data (2008). Data
file edition 3.0. Norwegian Social Science Data Services, Norway \textendash{}
Data Archive and distributor of ESS data. http://ess.nsd.uib.no/.%
} consists of answers from a survey given to the general population
of $28$ European countries. There are ten questions in the survey
where a valid answer is a number in the set $\left\{ 1,\ldots,10\right\} $.
The value $10$ represents the most positive opinion and $1$ the
most pessimistic opinion (we denote the name of the attribute in parenthesis):
\begin{itemize}
\item trust in local parliament (\texttt{country\_GOV})
\item trust in local politicians (\texttt{politicians})
\item trust in EU Parliament (\texttt{EU\_GOV})
\item trust in United Nations (\texttt{UN})
\item trust in country\textquoteright{}s parliament (\texttt{country\_GOV})
\item how satisfied with life (\texttt{Life})
\item how satisfied with the national government (\texttt{National\_GOV})
\item immigration is bad or good (\texttt{Immigration})
\item the state of health services (\texttt{Health})
\item how happy are you (\texttt{happy})
\end{itemize}
After normalizing each component we represent each country as a fuzzy
set on a domain that consists of the ten attributes. Table \ref{tab:FuzzyMembershipValues}
displays the membership functions for each of the countries. Each
row in this table represents a membership function $m_{i}(x)$ of
the fuzzy set $C_{i}$ of country $i$. Based on this information
we compute the distance $d(C_{i},C_{j})$ between every possible pair
of countries $C_{i},C_{j}$ and obtain a distance matrix $D=\left[d_{i,j}\right]$,
$d_{i,j}:=\text{dist}\left(C_{i},C_{j}\right)$. We use $D$ as the
newly transformed version of the original data (Table \ref{tab:FuzzyMembershipValues})
and do data-clustering on it. The $i^{th}$ row of $D$ is a feature
vector representation of country $i$. We use the $k$-means clustering
procedure.

Figure \ref{fig:The-result-of} shows the result of the $k$-means
clustering where the horizontal axis displays the cluster number and
the vertical axis shows the distance of each point in a cluster to
the mean of the cluster. In order to interpret this result, let us
look at the fuzzy sets of some of the clusters. Figure \ref{fig:Fuzzy-sets-representation}
displays the fuzzy sets of cluster $\#1$. As seen, the country Spain
is considered similar to the rest of the countries in this cluster
although its \textquotedblleft{}happy\textquotedblright{} value is
almost complement to the rest of the countries. This follows from
the complement-property of our distance function (see section \ref{sec:Simple-Examples}). 

Figure \ref{fig:cluster2} displays the fuzzy sets of cluster $\#2$.
Ukraine seems to behave almost the opposite of Denmark (besides on
the attributes \texttt{EU\_Gov} where both have similar values). Ukraine
and Turkey have interesting behaviors: they take very similar values
for the attributes \texttt{country-GOV} up to \texttt{Life} and on
\texttt{UN} while on the rest of the attributes they are almost mutually
complement. Hence according to our distance they are considered close
(which is why they are placed in the same cluster).

Figure \ref{fig:Fuzzy-sets-representationCLUSTER3} shows the fuzzy
sets of several countries in custer $\#3$. Hungary and the Russian
Federation take very similar values and hence are close. Israel versus
Hungary or versus Russian Federation has a similar behavior on the
attributes \texttt{country\_GOV}, \texttt{EU\_GOV}, \texttt{happy},
\texttt{National\_GOV}, \texttt{politicians}, \texttt{UN}, while on
\texttt{Health}, \texttt{Immigration}, \texttt{Life} it has almost
the complement values. Hence, overall, our distance function renders
Israel as close to Hungary and Russia.

We also ran a clustering procedure which is a variant of the Kohonen
Self Organizing Map. The results that we obtained are very similar
to those obtained by the $k$-means procedure.

\section{Conclusion}

This paper introduces a new distance function $\text{dist}(A,B)$
for fuzzy sets $A,B$ based on their descriptive complexity. The distance
is shown to be a semi-metric that satisfies the triangle inequality.
In comparison to other existing distance-functions for fuzzy sets
this new metric is proportional to the additional amount of information
needed to describe fuzzy set $A$ when knowing fuzzy set $B$ or vice
versa. It thus has a natural information-based interpretation. Doing
pattern clustering based on this distance we have shown that fuzzy
sets that are clustered together tend to be more mutually informative.
This is an interesting new property that can be useful for analyzing
other data sets.

\bibliographystyle{plainnat}
%\bibliography{/Users/Joel/MyStuff/MyBib/jerbib/jer}

\section*{Figures and Tables}

\begin{table}[h!] 	\centering 		\begin{tabular}{lcccccccccc} 			Belgium & 	0.60	& 0.55 &	0.69	& 0.60	& 0.60	& 0.72	& 0.47	& 0.48	& 1.00	& 0.30\\ 			Bulgaria &	0.06	& 0.41	& 0.99	& 0.45	& 0.06	& 0.06	& 0.59	& 0.61	& 0.15	& 0.04\\		   Switzerland 	& 0.85 &	0.63 &	0.66 &	0.65 &	0.85 &	0.86 &	0.79 &	1.00 &	0.90 &	0.49\\ 			  Cyprus 	& 0.78 &	0.79 &	0.96 &	0.43 &	0.78 &	0.66 &	0.91 &	0.37 &	0.67 &	0.37 \\ 			Czech R. &	0.32 &	0.63 &	0.68 &	0.47 &	0.32 &	0.56 &	0.69 &	0.24 &	0.59 &	0.26\\ 				Germany &	0.60 &	0.65 &	0.86 &	0.52 &	0.60 &	0.62 &	0.74 &	0.61 &	0.45 &	0.17\\ 			Denmark 	&	1.00 &	0.83 & 	0.91 &	0.92 &	1.00	& 1.00 &	0.81 &	0.67 &	0.68 &	1.00\\ 			Estonia 	& 0.46 &	0.41 &	0.91 &	0.60 &	0.46 &	0.48 &	0.84 &	0.37 &	0.53 &	0.23\\ 			Spain 	& 	0.68 & 	0.51 &	0.74 &	0.52 &	0.68 &	0.72 &	0.70 &	0.61 &	0.73 &	0.94\\ 		Finland 	& 	0.89 &	0.65 &	0.66 &	0.96 &	0.89 &	0.87 &	0.74 &	0.73 &	0.85 &	0.42\\ 		France 			& 0.58 &	0.39 &	0.72 &	0.57 &	0.58 &	0.49 &	0.59 &	0.50 &	0.71 &	0.25\\ 			UK 				& 0.54 &	0.62 &	0.53 &	0.52 &	0.54 &	0.66 &	0.75 &	0.39 &	0.71 &	0.28\\ 				Greece 	& 0.40 &	0.49 &	0.87 &	0.26 &	0.40 &	0.43 &	0.69 &	0.00	& 0.17 &	0.15\\ 				Coatia 	& 0.28 &	0.51 &	0.84 &	0.26 &	0.28 &	0.53 &	0.68 &	0.29 &	0.40 &	0.12\\ 			Hungary 	& 0.20 &	0.51 &	0.89 &	0.42 &	0.20 &	0.28 &	0.69 &	0.02 &	0.27 &	0.06\\ 			Israel 		&	0.45 &	0.69 &	0.86 &	0.22 &	0.45 &	0.75 &	0.80 &	0.67 &	0.78 &	0.19\\ 			Latvia 		&	0.07 &	0.19 &	0.75 &	0.40 &	0.07 &	0.41 &	0.28 &	0.20 &	0.22 &	0.12\\ 	Netherlands 	& 0.79 &	0.40 &	0.00	& 0.67 &	0.79 &	0.80 &	0.65 &	0.68 &	0.75 &	0.57\\ 			Norway 		&	0.85 &	0.64 &	0.75 &	1.00	&	0.85 &	0.86 &	0.72 &	0.83 &	0.72 &	0.93\\ 		Poland 			& 0.28 &	0.00 &	0.77 &	0.56 &	0.28 &	0.63 &	0.60 &	0.74 &	0.27 &	0.15\\ 		Portugal 		& 0.39 &	0.51 &	0.81 &	0.50 &	0.39 &	0.34 &	0.72 &	0.58 &	0.38 &	0.16\\ 		Romania 		& 0.45 &	0.75 &	1.00 & 	0.67 &	0.45 &	0.44 &	0.77 &	0.66 &	0.30 &	0.08\\ 		Russian Fed &	0.49 &	0.79 &	0.95 &	0.29 &	0.49 &	0.29 &	0.92 &	0.19 &	0.22 &	0.08\\ 			Sweden 		&	0.84 &	0.70 &	0.64 &	0.90	&	0.84 &	0.85 &	0.75 &	0.73 &	0.73 &	0.47\\ 			Slovenia 	&	0.57 &	0.74 &	0.92 &	0.51 &	0.57	&	0.64 &	0.90 &	0.33 &	0.48 &	0.24\\ 			Slovakia 	& 0.52 &	0.69 &	0.95 &	0.61 &	0.52 &	0.52 &	0.84 &	0.30 &	0.38 &	0.19 \\ 				Turkey 	&	0.87 &	1.00 & 	0.99 &	0.00	& 0.87 &	0.33 &	1.00	& 0.05 &	0.60 &	0.00\\ 			Ukraine 	& 0.00	& 0.38  &	0.94 &	0.08 &	0.00 &	0.00 &	0.00 &	0.31 &	0.00 &	0.05\\
		\end{tabular} 
\vskip .3cm
	\caption{Fuzzy membership values} 	\label{tab:FuzzyMembershipValues} \end{table} 

\begin{figure}[h]
\begin{raggedleft}
\includegraphics[scale=0.6]{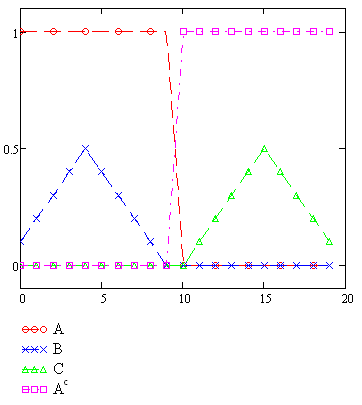}
\par\end{raggedleft}

\vskip -0.1cm\hspace{0.5in}\parbox{6in}{ \caption{\label{fig:Fuzzy-sets-,,}Fuzzy sets $A$,$B$,$C$ and $A^{c}$}

}
\end{figure}

\begin{figure}[h]
\begin{raggedleft}
\includegraphics[scale=0.6]{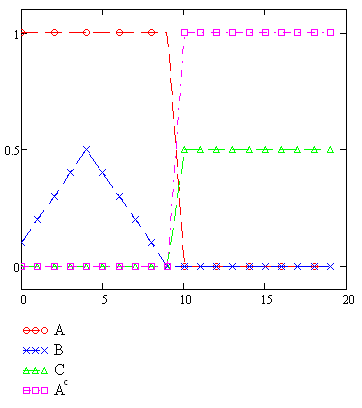}
\par\end{raggedleft}

\vskip -0.1cm\hspace{0.5in}\parbox{6in}{ \caption{\label{fig:Fuzzy-sets-,,-1}Fuzzy sets $A$,$B$,$C$ and $A^{c}$}

}
\end{figure}

\noindent \begin{flushleft}
\begin{figure}[h]
\begin{raggedleft}
\includegraphics[scale=0.6]{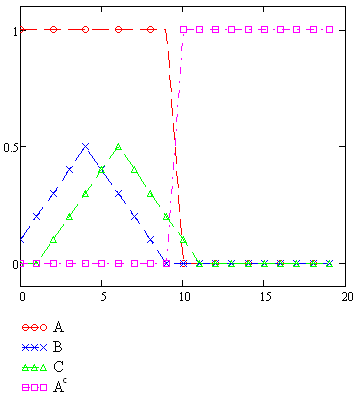}
\par\end{raggedleft}

\vskip -0.1cm\hspace{0.5in}\parbox{6in}{ \caption{\label{fig:Fuzzy-sets-,,-1-1}Fuzzy sets $A$,$B$,$C$ and $A^{c}$}

}
\end{figure}

\par\end{flushleft}

\newpage
\setlength\textwidth{8in} \setlength{\oddsidemargin}{-1cm} \setlength{\evensidemargin}{0cm} 

\noindent \begin{center}
\begin{figure}[h]
\begin{centering}
\includegraphics[width=7in]{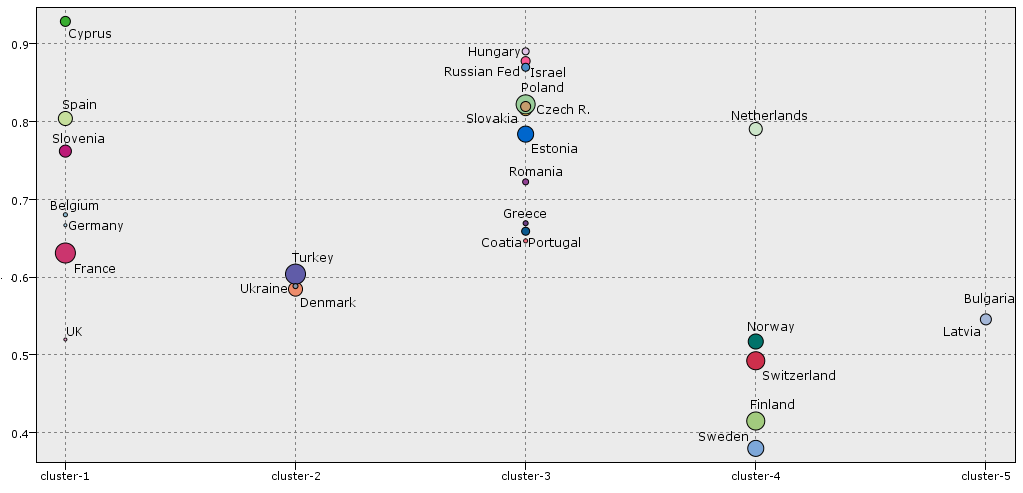}
\par\end{centering}

\vskip -0.1cm\hspace{0.5in}\parbox{6in}{ \caption{\label{fig:The-result-of}The result of $k$-means clustering of countries
based on the distance matrix $M$. The horizontal axis displays the
cluster number (there are five clusters). The vertical axis shows
the distance of each point in a cluster to the mean of the cluster. }

}
\end{figure}

\par\end{center}

\noindent \begin{flushleft}
\begin{figure}[h]
\begin{centering}
\includegraphics[width=7in]{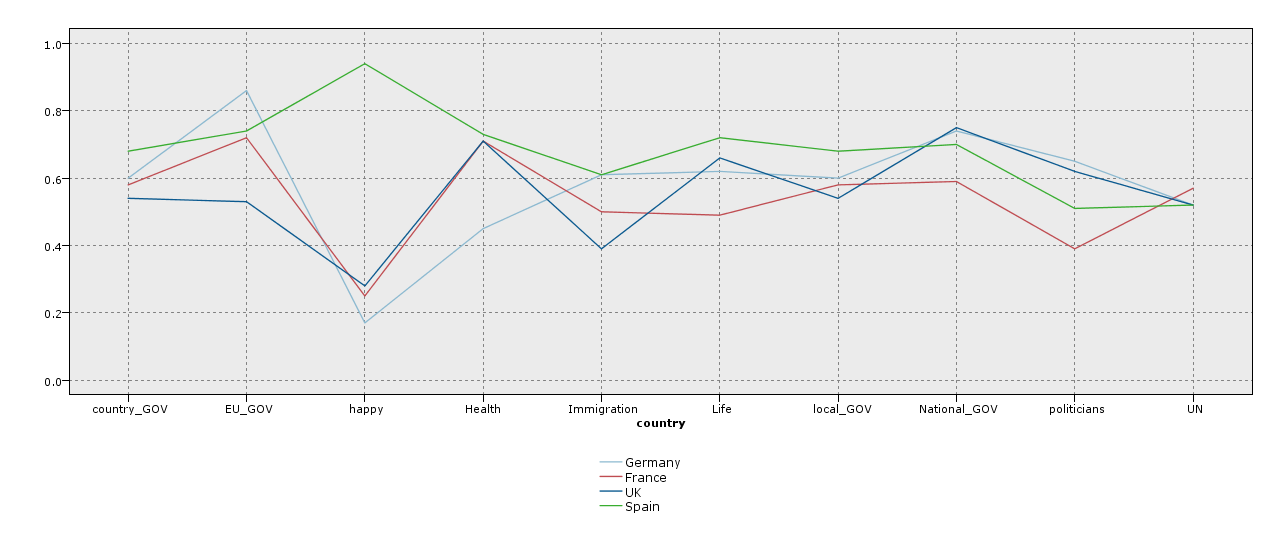}
\par\end{centering}

\vskip -0.5cm\hspace{0.5in}\parbox{6in}{ \caption{\label{fig:Fuzzy-sets-representation}Fuzzy sets representation of
the countries in Cluster $\#1$.}

}
\end{figure}

\par\end{flushleft}

\noindent \begin{flushleft}
\begin{figure}[h]
\noindent \begin{centering}
\includegraphics[width=7in]{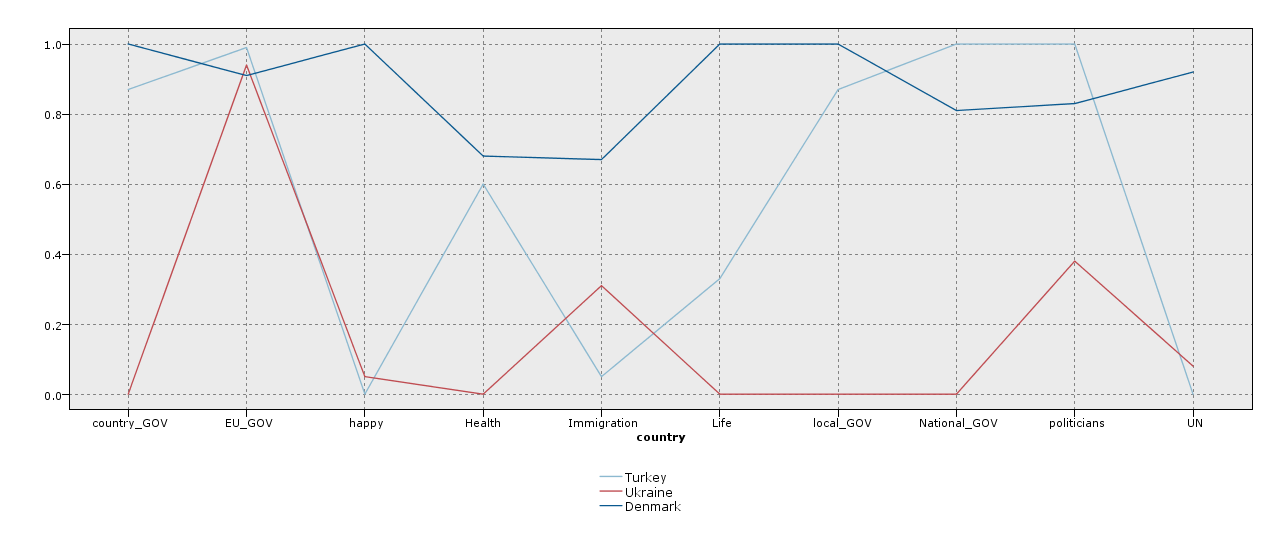}
\par\end{centering}

\vskip -0.5cm\hspace{0.5in}\parbox{6in}{ \caption{\label{fig:cluster2}Fuzzy sets representation of the countries in
Cluster $\#2$ }

}
\end{figure}

\par\end{flushleft}

\noindent \begin{flushleft}
\begin{figure}[h]
\begin{centering}
\includegraphics[width=7in]{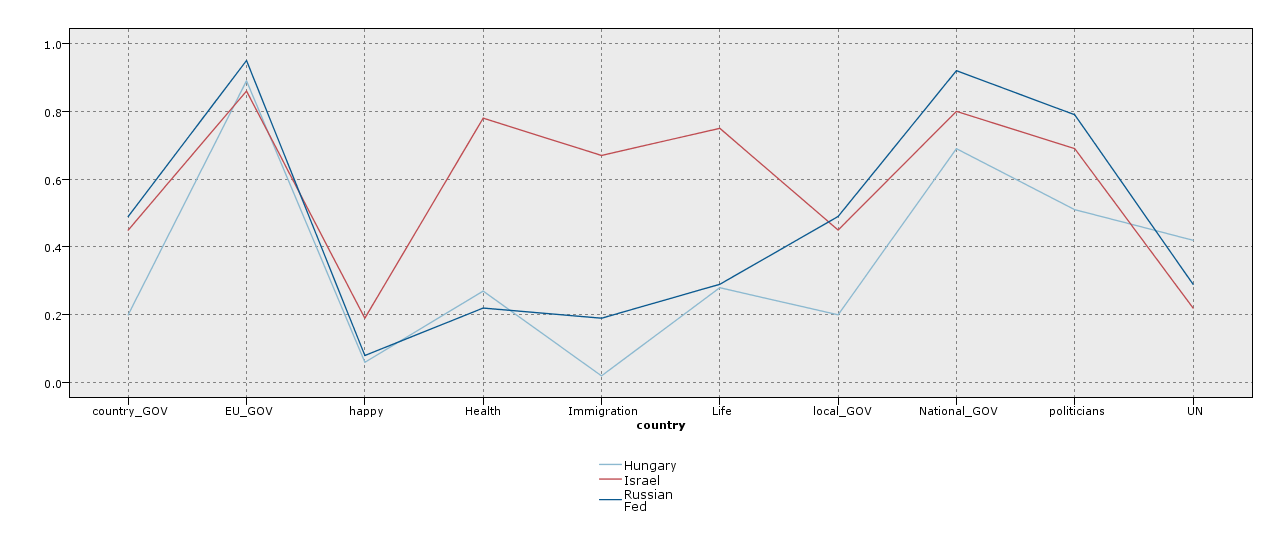}
\par\end{centering}

\vskip -0.5cm\hspace{0.5in}\parbox{6in}{ \caption{\label{fig:Fuzzy-sets-representationCLUSTER3}Fuzzy sets representation
of some of the countries in Cluster $\#3$.}
}
\end{figure}

\par\end{flushleft}
\end{document}